\documentclass[11pt,a4paper]{article}
\usepackage[hyperref]{emnlp2019}
\usepackage{times}
\usepackage{latexsym}
\usepackage{graphicx}
\usepackage{url}
\usepackage{booktabs}
\usepackage{multirow}
\usepackage{subcaption}
\usepackage{amsmath}
\usepackage{amsfonts}

\aclfinalcopy 

\title{Text Level Graph Neural Network for Text Classification}

\author{Lianzhe Huang, Dehong Ma, Sujian Li, Xiaodong Zhang and Houfeng WANG\\
 MOE Key Lab of Computational Linguistics, Peking University, Beijing, 100871, China \\
  {\tt \{hlz, madehong, lisujian, zxdcs, wanghf\}@pku.edu.cn} }

\date{}

\begin{document}
\maketitle
\begin{abstract}
Recently, researches have explored the graph neural network (GNN) techniques on text classification, since GNN does well in handling complex structures and preserving global information. 
However, previous methods based on GNN are mainly faced with the practical problems of fixed corpus level graph structure which do not support online testing and high memory consumption. 
To tackle the problems, we propose a new GNN based model that builds graphs for each input text with global parameters sharing instead of a single graph for the whole corpus. This method removes the burden of dependence between an individual text and entire corpus which support online testing, but still preserve global information. Besides, we build graphs by much smaller windows in the text, which not only extract more local features but also significantly reduce the edge numbers as well as memory consumption. 
Experiments show that our model outperforms existing models on several text classification datasets even with consuming less memory.
  
\end{abstract}

\section{Introduction}
Text classification is a fundamental problem of natural language processing (NLP), which has lots of applications like SPAM detection, news filtering, and so on~\cite{jindal2007review, aggarwal2012survey}. The essential step for text classification is text representation learning. 

With the development of deep learning, neural networks like Convolutional Neural Networks (CNN)~\cite{kim2014convolutional} and Recurrent Neural Networks (RNN)~\cite{hochreiter1997long} have been employed for text representation.
Recently, a new kind of neural network named Graph Neural Network (GNN) has attracted wide attention~\cite{battaglia2018relational}. GNN was first proposed in~\cite{scarselli2009graph} and has been used in many tasks in NLP including text classification~\cite{defferrard2016convolutional}, sequence labeling~\cite{zhang2018sentence}, neural machine translation~\cite{bastings2017graph}, and relational reasoning~\cite{battaglia2016interaction}. \newcite{defferrard2016convolutional} first employed Graph Convolutional Neural Network (GCN) in text classification task and outperformed the traditional CNN models. Further, \newcite{yao2018graph} improved \newcite{defferrard2016convolutional}'s work by applying article nodes and weighted edges in the graph, and their model outperformed the state-of-the-art text classification methods.

However, these GNN-based models usually adopt the way of building one graph for the whole corpus, which causes the following problems in practice.
First, high memory consumption is required due to numerous edges. Because this kind of methods build a single graph for the whole corpus and use edges with fixed weights, which considerably limits the expression ability of edges, they have to use a large connection window to get a global representation. 
Second, it is difficult for this kind of models to conduct the online test, because the structure and parameters of their graph are dependent on the corpus and cannot be modified after training.

To address the above problems, we propose a new GNN based method for text classification. Instead of building a single corpus level graph, we produce a text level graph for each input text. For a text level graph, we connect word nodes within a reasonably small window in the text rather than directly fully connect all the word nodes. The representations of the same nodes and weights of edges are shared globally and can be updated in the text level graphs through a massage passing mechanism, where a node takes in the information from neighboring nodes to update its representation. Finally, we summarize the representations of all the nodes in the graph to predict the results. 
With our design, text level graphs remove the burden of dependency between a single input text and the entire corpus, which support online test. 
Besides,  
it has the benefit of consuming less memory by connecting words in a small contextual window, because it excludes a good many words that are far away in the text and have little relation with the current word and thus significantly reduces the number of edges.
The message passing mechanism makes nodes in the graph perceive information around them to get precise meaning in a specific context.

In our experiments, our method achieves state-of-the-art results in several text classification datasets and consumes significantly fewer memory resources compared with previous methods.

\section{Method}

\begin{figure}[t]
    \centering
    \includegraphics[width=0.45\textwidth]{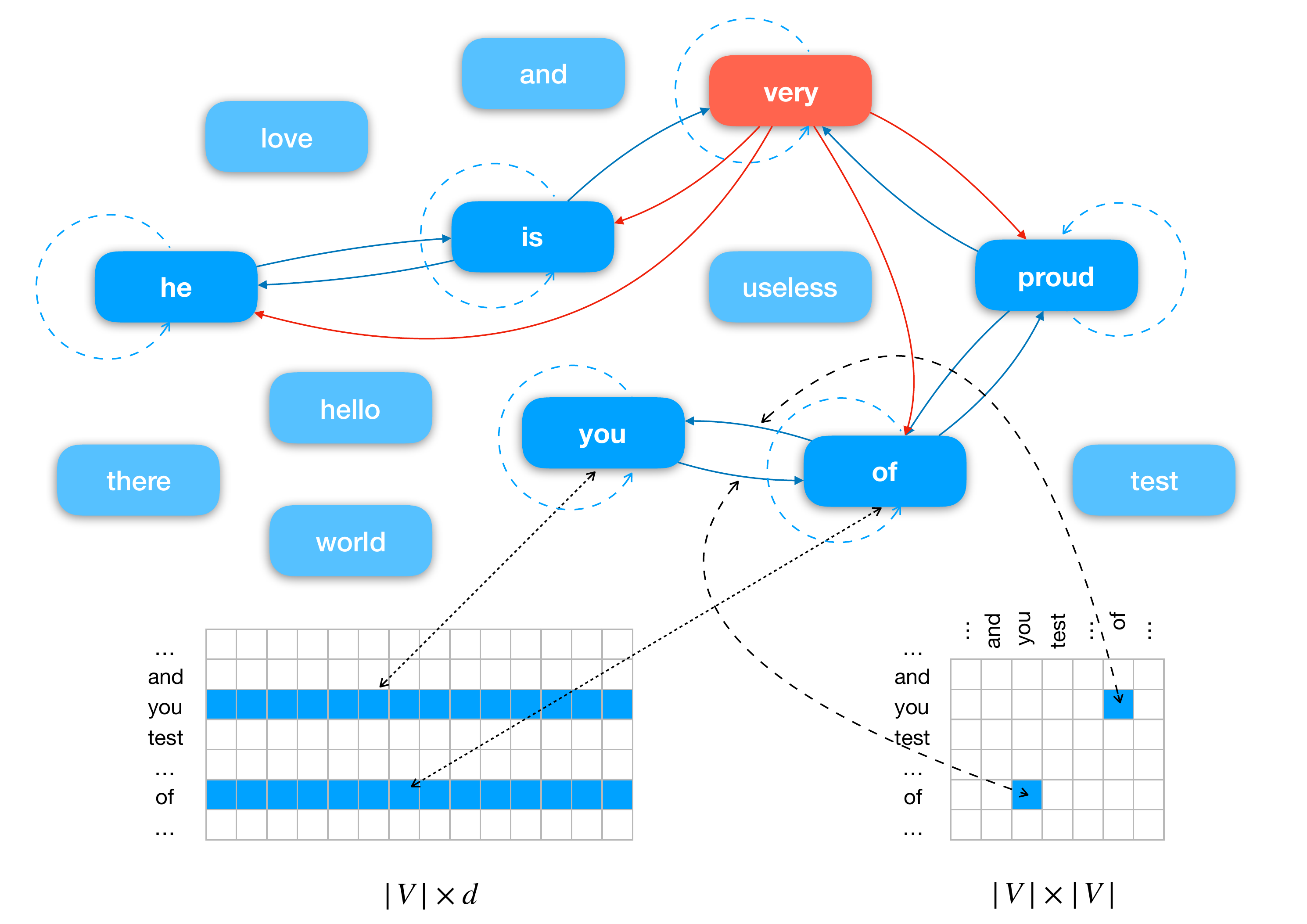}
    \caption{Structure of graph for a single text ``he is very proud of you.". For the convenience of display, in this figure, we set $p=2$ for the node ``very" (nodes and edges are colored in red) and  $p=1$ for the other nodes(colored in blued). In actual situations, the value of $p$ during a session is unique. All the parameters in the graph come from the global shared representation matrix, which is shown at the bottom of the figure.}
    \label{fig:main_flg}
\end{figure}

In this section, we will introduce our method in detail. 
First, we show how to build a text level graph for a given text; all the parameters for the text level graph are taken from some global-sharing matrices. 
Then, we introduce the message passing mechanism on these graphs to obtain information from the context. 
Finally, we depict how to predict the label for a given text based on the learned representations.
The overall architecture of our model is shown in Figure \ref{fig:main_flg}.

\subsection{Building Text Graph}\label{graph_def}

We notate a text with $l$ words as $T=\{\mathbf{r_1},...\mathbf{r_i},...,\mathbf{r_l}\}$, where $\mathbf{r_i}$ denotes the representation of  the $i_{th}$ word.
$\mathbf{r_i}$ is a vector initialized by $d$ dimension word embedding and can be updated by training. To build a graph for a given text, we regard all the words that appeared in the text as the nodes of the graph. Each edge starts from a word in the text and ends with its adjacent words. Concretely, the graph of text $T$ is defined as:
\begin{align}
    N &= \{\mathbf{r_i}| i\in[1, l]  \},& \\
    E &= \{e_{ij}| i\in[1, l]; j[i-p, i+p]\},
\end{align}
where $N$ and $E$ are the node set and edge set of the graph, and word representations in $N$ and edge weights in $E$ are taken from global shared matrices. $p$ denotes the number of adjacent words connected to each word in the graph. Besides, we uniformly map the edges that occur less than $k$ times in the training set to a ``public" edge to make parameters adequately trained.

Compared with the previous methods in building graph, our approach can exceedingly reduce the scale of the graph in terms of nodes and edges. That means that the text-level graph can consume much less GPU memory. Besides, their method is unfriendly to new-coming text, while our approach can solve this problem because the graph for each text is only dependent on its content.

\subsection{Message Passing Mechanism}\label{sec:mpm}

Convolution can extract information from local features~\cite{lecun1989backpropagation}. In the graph domain, convolution is implemented by spectral approaches~\cite{bruna2013spectral, henaff2015deep}, or non-spectral approaches~\cite{duvenaud2015convolutional}. In this paper, a non-spectral method named message passing mechanism~(MPM)~\cite{gilmer2017neural} is employed for convolution. MPM first collects information from adjacent nodes and updates its representations based on its original representations and collected information, which is defined as:
\begin{align}
\mathbf{M_n} &=  \max \limits_{a\in \mathcal{N}_n^p} e_{an}\mathbf{r_a} \label{message}, \\
\mathbf{r_n^{\prime}} &= (1-\eta_{n})\mathbf{M_n} + \eta_{n}\mathbf{r_n} \label{update}
\end{align}
where $\mathbf{M_n}\in \mathbb{R}^{d}$ is the messages that node $n$ receives from its neighbors; $\max$ is a reduction function which combines the maximum values on each dimension to form a new vector as an output.  $\mathcal{N}_n^p$ denotes nodes that represent the nearest $p$ words of $n$ in the original text; $e_{an} \in \mathbb{R}^1$ is the edge weight from node $a$ to node $n$, and it can be updated during training; and $\mathbf{r_n} \in \mathbb{R}^{d}$ denotes the former representation of node $n$.  $\eta_n \in \mathbb{R}^{1}$ is a trainable variable for node $n$ that indicates how much information of  $\mathbf{r_n}$ should be kept. $\mathbf{r_n^{\prime}}$ denotes the updated representation of node $n$.

MPM makes the representations of nodes influenced by neighborhoods, which means the representations can bring the information from context. Therefore, even for polysemous words, the precise meaning in the context can be determined by the influence of weighted information from neighbors. Besides, the parameters of text level graphs are taken from global shared matrices, which means the representations can also bring global information as other graph-based models do. 

Finally, the representations of all nodes in the text are used to predict the label of the text:
\begin{align}
    y_i = \textrm{softmax}(\textrm{Relu}(\mathbf{W}\sum_{n\in N_i}\mathbf{r_n^{\prime}} + \mathbf{b}))
\end{align}
where $W\in \mathbb{R}^{d\times c}$ is a matrix mapping the vector into an output space, $N_i$ is the node set of text $i$ and $\mathbf{b}\in \mathbb{R}^{c}$ is bias. 

The goal of training is to minimize the cross-entropy loss between ground truth label and predicted label:
\begin{align}
    \mathbf{loss} &= -g_i\log y_i,
\end{align}
where $g_i$ is the ``one-hot vector'' of ground truth label.

\section{Experiments}\label{sec:exp_setup}
In this section, we describe our experimental setup and report our experimental results.
\subsection{Experimental Setup}

For experiments, we utilize datasets including R8, R52\footnote{https://www.cs.umb.edu/\~{}smimarog/textmining/datasets/}, and  Ohsumed\footnote{http://disi.unitn.it/moschitti/corpora.htm}.
R8 and R52 are both the subsets of Reuters 21578 datasets. Ohsumed corpus is extracted from MEDLINE database. MEDLINE is designed for multi-label classification, we remove the text with two or more labels. For all the datasets above, we randomly select 10\% text from the training set to build validation set.  The overview of datasets is listed in Table \ref{tab:datasets}.

We compare our method with the following baseline models. 
It is noted that the results of some models are directly taken from~\cite{yao2018graph}.

\begin{itemize}
     \item \textbf{CNN} Proposed by~\cite{kim2014convolutional}, perform convolution and max pooling operation on word embeddings to get representation of text. 
     
     \item \textbf{LSTM} Defined in~\cite{liu2016recurrent}, use the last hidden state as the representation of the text. Bi-LSTM is a bi-directional LSTM.
     
     \item \textbf{fastText} Proposed by~\cite{joulin2016bag}, average word or n-gram embeddings as documents embeddings.
     
    \item \textbf{Graph-CNN} Operate convolution over word embedding similarity graphs by fourier filter, proposed by~\cite{defferrard2016convolutional}.
    
    \item \textbf{Text-GCN} A graph based text classification model proposed by~\cite{yao2018graph}, which builds a single large graph for whole corpus.
\end{itemize}

\subsection{Implementation Details}
We set the dimension of node representation as 300 and initialize with random vectors or Glove~\cite{pennington2014glove}. $k$ discussed in Section \ref{graph_def} is set to 2. We use the Adam optimizer~\cite{kingma2014adam} with an initial learning rate of $10^{-3}$, and L2 weight decay is set to $10^{-4}$.  Dropout with a keep probability of 0.5 is applied after the dense layer. The batch size of our model is 32. We stop training if the validation loss does not decrease for 10 consecutive epochs. 

For baseline models, we use default parameter settings as in their original papers or implementations. For models using pre-trained word embeddings, we used 300-dimensional GloVe word embeddings.

\subsection{Experimental Results}

Table \ref{acc-table} reports the results of our models against other baseline methods.
We can see that our model can achieve the state-of-the-art result. 
\begin{table}[t]
\centering
\footnotesize
\scalebox{1}{
\begin{tabular}{@{}ccccc@{}}
\toprule
\textbf{Datasets} & \textbf{\# Train} & \textbf{\# Test} & \textbf{Categories} & \textbf{Avg. Length} \\ \midrule
R8                & 5485              & 2189             & 8                   & 65.72                   \\
R52               & 6532              & 2568             & 52                  & 69.82                   \\
Ohsumed           & 3357              & 4043             & 23                  & 135.82                  \\ \bottomrule
\end{tabular}
}
\caption{\label{tab:datasets} Datasets overview.}
\end{table}

\begin{table}[tb]
    \centering
    \footnotesize
    \scalebox{1}{
    \begin{tabular}{cccc}
        \toprule
        \textbf{Model} & \textbf{R8}              & \textbf{R52}            & \textbf{Ohsumed}        \\ \midrule
        CNN            & 94.0 $\pm$ 0.5          & 85.3 $\pm$ {0.5}          & 43.9 $\pm$ {1.0}          \\
        LSTM           & 93.7 $\pm$ 0.8          & 85.6 $\pm$ {1.0}          & 41.1 $\pm$ {1.0}          \\
        Graph-CNN      & 97.0 $\pm$ 0.2          & 92.8 $\pm$ {0.2}          & 63.9 $\pm$ {0.5}        \\
        Text-GCN       & 97.1 $\pm$ 0.1          & 93.6 $\pm$ {0.2}          & 68.4 $\pm$ 0.6 \\
         \midrule
        CNN*           & 95.7 $\pm$ {0.5}          & 87.6 $\pm$ {0.5}          & 58.4 $\pm$ {1.0}          \\
        LSTM*          & 96.1 $\pm$ {0.2}          & 90.5 $\pm$ {0.8}          & 51.1 $\pm$ {1.5}          \\
        Bi-LSTM*       & 96.3 $\pm$ {0.3}          & 90.5 $\pm$ {0.9}          & 49.3 $\pm$ {1.0}          \\
        fastText*      & 96.1 $\pm$ {0.2}          & 92.8 $\pm$ {0.1}          & 57.7 $\pm$ {0.5}          \\
        Text-GCN*      & 97.0 $\pm$ {0.1}          & 93.7 $\pm$ {0.1}          & 67.7 $\pm$ {0.3}          \\
        Our Model*     & \textbf{97.8} $\pm$ \textbf{0.2} & \textbf{94.6} $\pm$ \textbf{0.3} & \textbf{69.4} $\pm$ \textbf{0.6} \\ \bottomrule
    \end{tabular}
    }
    \caption{\label{acc-table} Accuracy on several text classification datasets. Model with "*" means that all word vectors are initialized by Glove word embeddings. We run all models 10 times and report mean results. }
\end{table}

We note that the results of graph-based models are better than traditional models like CNN, LSTM, and fastTest. That is likely due to the characteristics of the graph structure. Graph structure allows a different number of neighbor nodes to exist, which enables word nodes to learn more accurate representations through different collocations. Besides, the relationship between words can be recorded in the edge weights and shared globally. These are all impossible for traditional models. 

We also find that our model performs better than graph-based models like Graph-CNN. Graph-CNN represents documents using the bag-of-word model, which is similar to ours, but they connect word nodes within a large window without weighted edges, which cannot distinguish the importance between different words. While our model employed trainable edge weights, which let words express themselves differently when faced with various collocation. Besides, the weights are shared globally which means they can be trained by all the text contains the same collocation in the entire corpus. 

We also note that our model performs better than former state-of-the-art model Text-GCN.
That is likely due to more expressive edges, which have been discussed before, and the difference of representations learning. Text-GCN learns word representations by corpus level co-occurrence while our model is trained within a contextual window like traditional word embeddings. 
Therefore our model can benefit from pre-trained word embeddings and achieve better results.

\subsection{Analysis of Memory Consumption}

\begin{table}[t]
\centering
\footnotesize
\scalebox{1}{
\begin{tabular}{@{}ccc@{}}
\toprule
\textbf{Datasets} & \textbf{Text-GCN}  & \textbf{Our Model} \\ \midrule
R8      & 9,979M(2,841,760)  & 954M(250,623)      \\
R52      & 8,699M(3,574,162)  & 951M(316,669)      \\
Ohsumed  & 13,510M(6,867,490) & 1,167M(419,583)    \\ \bottomrule
\end{tabular}
}
\caption{\label{mem-table} Comparison of memory consuming. The number of edges in the whole model is in parentheses.}
\end{table}

Table \ref{mem-table} reports the comparison of memory consumption and edges numbers between Text-GCN and our model. Results show that our model has a significant advantage in memory consumption. 

As discussed in \ref{graph_def}, the words in our model are only connected to adjacent words in the texts, while Text-GCN , which is based on the corpus level graph, connects nodes within a reasonably large window. Because Text-GCN uses co-occurrence information as fixed weights, it has to enlarge the window size to get a more accurate co-occurrence weight. Therefore, we will get a much more sparse edge weights matrix than Text-GCN. Also, since the representation of a text is calculated by the sum of the representations of word nodes in the text, there is no text node in our model, which also reduces memory consumption.

\subsection{Analysis of Edges}
\begin{figure}[t]
    \centering
    \includegraphics[width=0.5\textwidth]{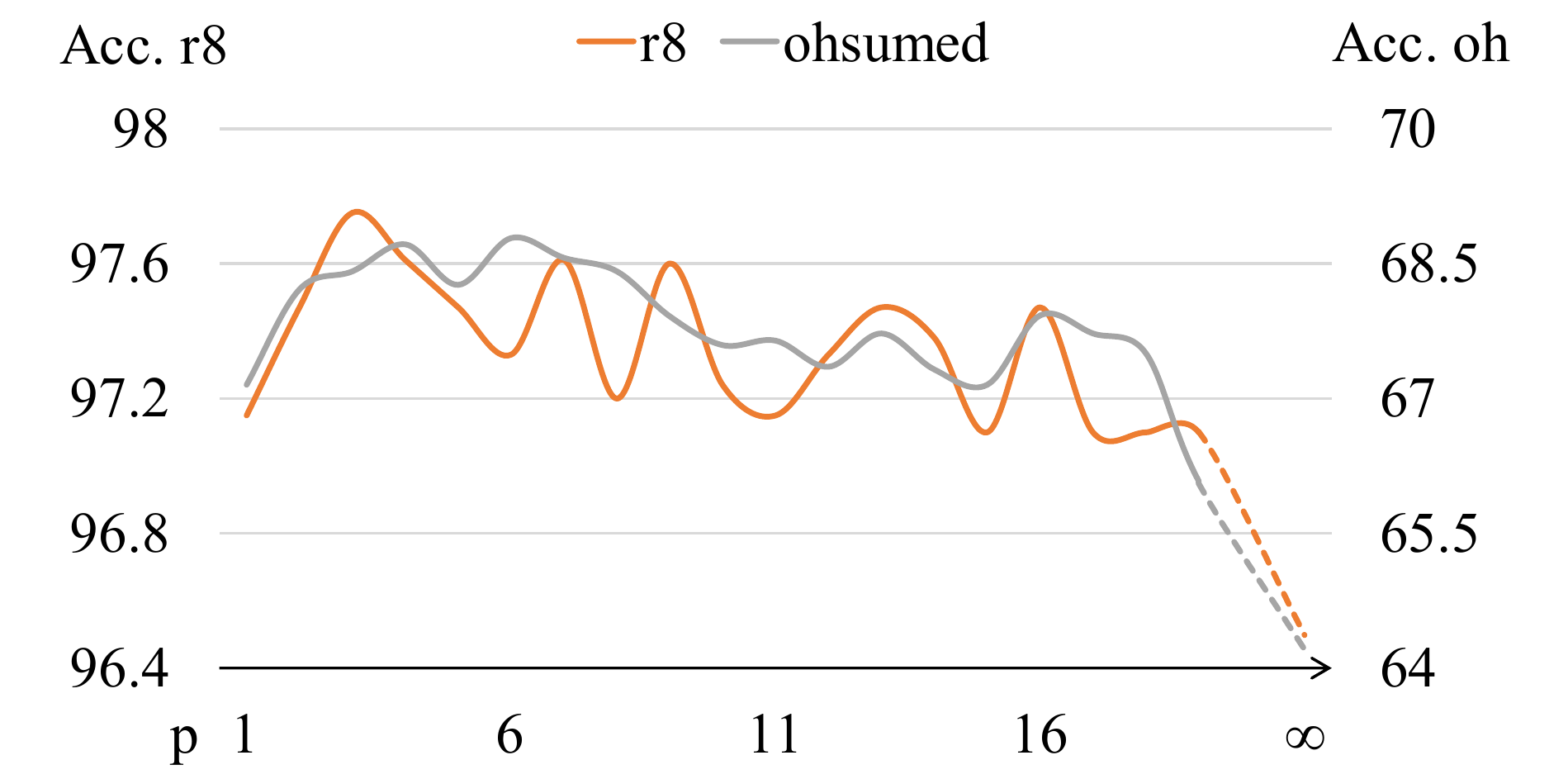}
    \caption{Model performance using $p$ from 1 to 19 and ``$\infty$'' (fully-connected). All hyperparameters are set the same except $p$. The left and right ordinate indicate the accuracy on the r8 and ohsumed dataset respectively.}
    \label{fig:p}
\end{figure}

To understand the difference of various connecting windows, we compared the performance of the R8 and ohsumed datasets with different $p$ values, the result is reported in Figure \ref{fig:p}. We find that the accuracy increases as $p$ becomes larger and achieves the best performance when connected with about 3 neighborhoods. Then the accuracy decreases volatility as $p$ increases. This suggests that when connected only with the nearest neighborhood, nodes cannot understand the dependencies that span multiple words in the context, while connected with neighborhoods far away (much larger $p$), the graphs become more and more similar with fully connected graphs which ignore the local features. 
In addition, the fewer edges, the fewer memory consumption. Our model has fewer edges compared with previous methods, and this also show the advantages of our proposed model.


\subsection{Ablation Study}
To further analyze our model, we perform ablation studies and Table \ref{abstudy} shows the results.
\begin{table}[t]
\centering
\footnotesize
\scalebox{0.9}{
\begin{tabular}{@{}lccc@{}}
\toprule
\textbf{Setting}  & \textbf{R8}           & \textbf{R52}          &\textbf{Ohsumed}      \\ \midrule
Original         & \textbf{97.8 $\pm$ 0.2} & \textbf{94.6 $\pm$ 0.3} & \textbf{69.4 $\pm$ 0.6} \\
(1)Fixed PMI Edges W. & 97.7 $\pm$ 0.2 & 94.0 $\pm$ 0.2 & 67.6 $\pm$ 0.5 \\ 
(2)Mean Reduction                            & 97.7 $\pm$ 0.1       & 94.5 $\pm$ 0.3       & 62.6 $\pm$ 0.2\\
(3)Random Word Emb. & 97.4 $\pm$ 0.2 & 93.7 $\pm$ 0.2 & 67.3 $\pm$ {0.5}          \\
\bottomrule
\end{tabular}
}
\caption{\label{abstudy} Results of ablation studies. We run all models for 5 times and give mean results. }

\end{table}

In (1), we fix the weights of edges and initialize them with point-wise mutual information (PMI), and the size of sliding windows is set to 20, which is the same as~\cite{yao2018graph}. Removing the trainable edges makes the model perform worse on all data sets, which demonstrates the effectiveness of trainable edges. 
In our opinion, the main reason is that trainable edges can better model the relations between words compared with fixed edges.

In (2), we change the max-reduction by mean-reduction. In the original model, the node gets its new representation from received messages by obtaining the maximum value alone each dimension. From Table \ref{abstudy}, we can see that the $\max$ reduction can achieve better results. The node reduction function is similar to the pooling operation on CNN. Reduction by $\max$ highlights features that are highly discriminating and provides non-linearity, which helps to achieve better results.

In (3), we remove the pre-trained word embeddings from nodes and initialize all the nodes with random vectors. Compared with the original model, the performances are slightly decreased without pre-trained word embeddings. Therefore, we believe that the pre-trained word embeddings have a particular effect on improving the performance of our model.


\section{Related Work}
In this section, we will introduce the related works about GNN and text classification in detail. 
\subsection{Graph Neural Networks}
Graph Neural Networks (GNN) has got extensive attention recently~\cite{zhou2018graph, zhang2018deep, wu2019comprehensive}. GNN can model non-Euclidean data, while traditional neural networks can only model regular grid data. While many tasks in reality such as knowledge graphs~\cite{hamaguchi2017knowledge}, social networks~\cite{hamilton2017inductive} and many other research areas~\cite{khalil2017learning} are with data in the form of trees or graphs. So GNN are proposed~\cite{scarselli2009graph} to apply deep learning techniques to data in graph domain. 

\subsection{Text Classification}
Text classification is a classic problem of natural language processing and has a wide range of applications in reality. Traditional text classification like bag-of-words~\cite{zhang2010understanding}, n-gram~\cite{wang2012baselines} and Topic Model~\cite{wallach2006topic} mainly focus on feature engineering and algorithms. With the development of deep learning techniques, more and more deep learning models are applied for text classification. \newcite{kim2014convolutional, liu2016recurrent} applied CNN and RNN into text classification and achieved results which are much better than traditional models. 

With the development of GNN, some graph-based classification models are gradually emerging~\cite{hamilton2017inductive, velivckovic2017graph, peng2018large}. \newcite{yao2018graph} proposed Text-GCN and achieved state-of-the-art results on several mainstream datasets. However, Text-GCN has the disadvantages of high memory consumption and lack of support online training. The model presents in this paper solves the mentioned problems in Text-GCN and achieves better results.

\section{Conclusion}
In this paper, we proposed a new graph based text classification model, which uses text level graphs instead of a single graph for the whole corpus. Experimental results show that our model achieves state-of-the-art performance and has a significant advantage in memory consumption. 

\section*{Acknowledgments}

Our work is supported by the National Key Research and Development Program of China under Grant No.2017YFB1002101 and National Natural Science Foundation of China under Grant No.61433015 and No.61572049. The corresponding author of this paper is Houfeng Wang. \\

\bibliography{emnlp2019}
\bibliographystyle{acl_natbib}

\appendix

\end{document}